\def\year{2021}\relax
\documentclass[letterpaper]{article} 
\usepackage{aaai21}  
\usepackage{times}  
\usepackage{helvet} 
\usepackage{courier}  
\usepackage[hyphens]{url}  
\usepackage{graphicx} 
\usepackage{natbib}  
\usepackage{caption} 
\usepackage{amssymb,amsmath,textcomp}
\title{An Analysis Of Protected Health Information Leakage\\ In Deep-Learning Based De-Identification Algorithms}
\author{
    Salman Seyedi, \textsuperscript{\rm 1} Li Xiong, \textsuperscript{\rm 2} Shamim Nemati, \textsuperscript{\rm 3} Gari D. Clifford \textsuperscript{\rm 1,4}\\
}
\affiliations{
    \textsuperscript{\rm 1}Department of Biomedical Informatics, Emory University, Atlanta, GA 30322, USA\\
    \textsuperscript{\rm 2}Department of Computer Science, Emory University, Atlanta, GA 30322, USA\\
    \textsuperscript{\rm 3}Department of Biomedical Informatics, University of California San Diego Health, La Jolla, CA 92037, USA\\
    \textsuperscript{\rm 4}Department of Biomedical Engineering, Georgia Institute of Technology, Atlanta, GA 30308, USA\\
    sseyedi@emory.edu, lxiong@emory.edu, snemati@ucsd.edu, gari@gtech.edu

}

\begin{document}

\maketitle

\begin{abstract}
The increasing complexity of algorithms for analyzing medical data, including de-identification tasks, raises the possibility that complex algorithms are learning not just the general representation of the problem, but specifics of given individuals within the data. Modern legal frameworks specifically prohibit the intentional or accidental distribution of patient data, but have not addressed this potential avenue for leakage of such protected health information. Modern deep learning algorithms have the highest potential of such leakage due to complexity of the models. Recent research in the field has highlighted such issues in non-medical data, but all analysis is likely to be data and algorithm specific. We, therefore, chose to analyze a state-of-the-art free-text de-identification algorithm based on LSTM (Long Short-Term Memory) and its potential in encoding any individual in the training set. Using the i2b2 Challenge Data, we trained, then analyzed the model to assess whether the output of the LSTM, before the compression layer of the classifier, could be used to estimate the membership of the training data. Furthermore, we used different attacks including membership inference attack method to attack the model. Results indicate that the attacks could not identify whether members of the training data were distinguishable from non-members based on the model output. This indicates that the model does not provide any strong evidence into the identification of the individuals in the training data set and there is not yet empirical evidence it is unsafe to distribute the model for general use.
\end{abstract}

\noindent An electronic health record, or electronic medical record (EMR) includes a wealth of information in the form of both physiological data and structured or free text. The latter is often replete with protected health information (PHI) and personal identifiable information (PII). As a result, there has been much attention paid to the notion of secure processing and sharing. The integrity of patients{\textquotesingle} personal information and related privacy are governed in the US by the Health Insurance Portability and Accountability Act of 1996 (HIPAA). Although intended to make medical data portable, it has had a much larger effect in ensuring privacy through de-identification of shared data.
The HIPAA privacy rules provide two avenues that one must follow to meet the de-identification standards: `expert determination' and `safe harbor'. The safe harbor de-identification method requires the removal of Protected Health Information (PHI) which is a list of 18 categories of identifiers. The process of de-identification of a particular document can be performed using different means, but based on the enormous amount of EMR which grows every moment, the sheer volume not only greatly incentivizes the automation of the process as much as possible, but one can argue any pragmatic approach have to rely intensively on utilizing power of computers. Because of this, the automation of the de-identification process has been of great interest and different algorithms have been developed over the years to facilitate the de-identification by automating the finding/labeling of the PHI \citep{Neamatullah2008,yang2015automatic,2016deidentification}.

In recent years due to improvements in the deep learning architectures and recurrent neural networks (RNN), not only there have been great achievements in improved metrics like accuracy, recall, and F1-score for the de-identification systems, but also the flexibility of the systems allow the entities to utilize the same algorithm on different corpora of EMRs. This property means that an entity can train the system on one corpus of records and then share that trained system to de-identify another body of reports, or even share it with other entities. This property, also known as transfer learning, while very useful and promising, raises the concern of leakage of sensitive information through these sharing processes \citep{melis2019exploiting}.

 One of the most widely used RNN is Long Short-Term Memory (LSTM) architecture, which is applied in many deep-learning based de-identification systems \citep{lample2016neural,kim2018ensemble,2016deidentification,shickel2017deep}. \\
One of these successful algorithms is NeuroNER \citep{2016deidentification}. LSTM provides a great capability for processing the long dependencies which is an important characteristic in text formats, thus, providing a very powerful tool. However, since it has a great number of variables (for instance more than 40000 for a single LSTM with 100 units), there is the concern of memorization of data and leakage of the sensitive information in the trained model due to the model complexity. All these extensive and expensive effort of the de-identification systems is to protect the privacy of the individuals. The leak of information through the parameters of the de-identifications model can jeopardize the privacy and nullify the main goal. 
One of the recent works on investigating the data memorization in neural network is 
\citep{carlini2019secret} on generative sequence models which is a specific neural network with LSTM. They study these particular LSTMs and show how memorization or leakage can happen even when the data is rare. Moreover, they suggest a quantitative metric for measuring the memorization on rare or unique sequences in the data. While they provide very elegant approach, their method is not directly applicable to neural networks other than generative ones.

In this article, we focus on the unintended memorization or leakage of the deep-learning de-identifiers and more specifically, the NeuroNER system. The reason behind selecting this system is the fact that it has been used in a few publicly accessible data sets with different structures ({\em i2b2 2014 challenge data set \citep{stubbs2015annotating}} and the MIMIC II de-identification data set \citep{Neamatullah2008}) and the same system with essentially same structure and hyperparameters, has been utilized to achieve the state-of-the-art performance \citep{2017neuroner} \citep{2016deidentification}. This approach provides a flexible framework and has the potential to be leveraged in transfer learning paradigms. As such, we are concerned that this type of approach can encode identities of the individuals in the training data into the weights of the neural network. In this work we explore this concept. First we provide a statistical analysis and perform cut-off attacks to determine the risk if this state-of-the-art algorithm has created unnecessary exposure for the data subjects. Moreover, we performed membership inference attack \citep{shokri2017membership}, which is the state-of-the-art re-identification attack related to our approach.

\section{materials and methods}\label{methods}
\subsection{NeuroNER}\label{NeuroNER}
In this subsection, a brief description of the NeuroNER package as well as specifics of how the system is trained are provided. More details can be found in 
\citep{2016deidentification,2017neuroner}. The statistical and inference attack methods are discussed afterward.
 
\begin{figure*}
    \centering
    \includegraphics[width=0.8\textwidth]{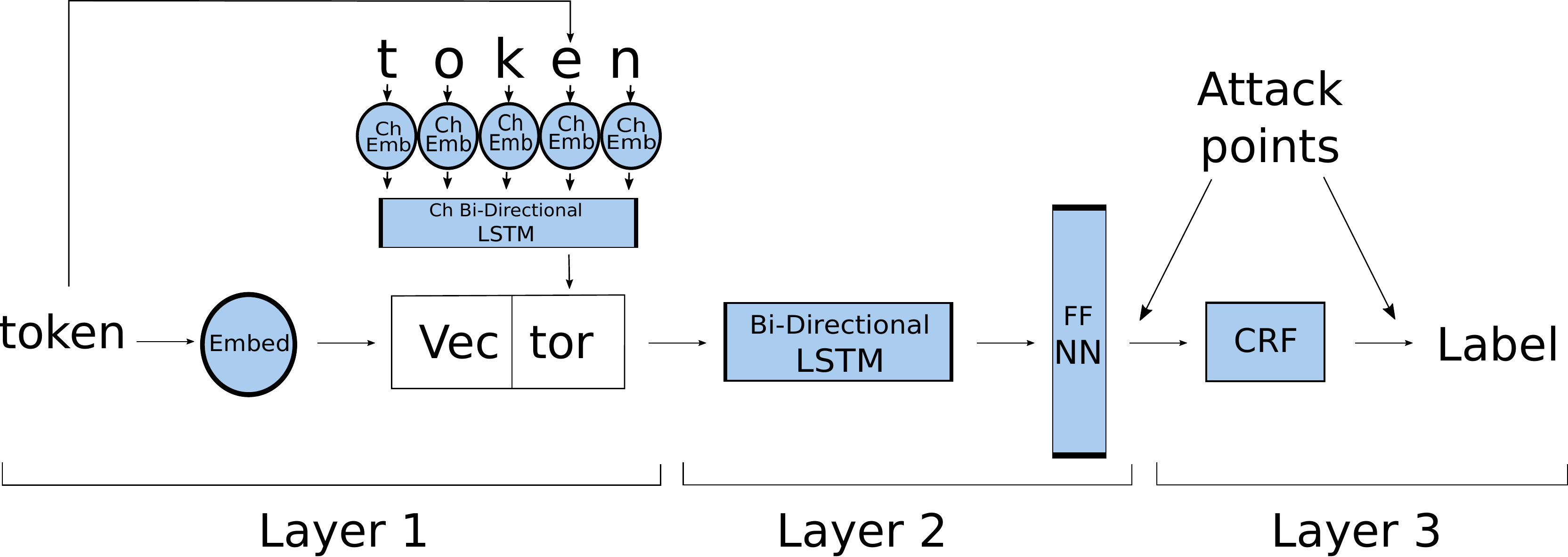}
    \caption{In this chart the circles represent character (Ch) embedding (Emb) and token level embedding (Embed), the FFNN indicates feed forward neural network. The attaching points are at the output level of the second layer after a feed-forward neural network and after the CRF in the third layer as indicated.}
    \label{fig:Flow}
\end{figure*}

 Briefly, the structure of the NeuroNER system, as can be seen in figure \ref{fig:Flow},
 is composed of three layers:
 \begin{itemize}
     \item Token embedding
     \item LSTM based label prediction
     \item CRF (conditional random field)
 \end{itemize}
 In the first layer, in parallel with a tokenizer and token embedding, there is an LSTM network for character level embedding. To avoid any memorization or leak, the token embedding was fixed (no learning) and loaded from the pre-trained models in public data (as is suggested in the original release codes). The output of the first layer, which is a concatenation of the word and character level embedding (illustrated as 'Vec|tor' in figure \ref{fig:Flow}), is sent to the bidirectional LSTM  of the second layer. The output of the LSTM is then sent to a feed-forward neural network which will output the probability vectors $P$. In the third layer, these probability vectors will be the input of the CRF. The CRF layer can be turned on or off as an option. 
 
 The first step in analyzing a text is to break it down to its compartments. In a sense, one can think of tokenization as separating each word in a text and calling it token. There are different ready-to-use tokenizers. In this work, we apply a method known as `Spacy', which is a free and open-source library for advance natural language processing (NLP).
 
 There are many different methods available to convert tokens into numbers. One simple approach would be representing any word with one-hot encoding. Given a vector that has the size of the word domain, the component corresponding to the word will have value one and all other components in the vector are zero (so naming is one-hot). In this approach, all the words are perpendicular to one another and the size of the space grows with the number of words. A more advanced approach is to use a lower-dimensional (100 for instance) denser space when the words are represented with vectors that are not all perpendicular and related words have smaller angles with each other. This representation is built by an unsupervised learning algorithm that leverages a large corpus of data, such as Wikipedia, and learns the relevance of the words in texts (mostly by their co-appearance in the text). In this paper, for the token embedding, a pre-trained embedding known as Global Vectors for Word Representation (GloVe) \citep{pennington2014glove} was used which is of the later type described above. 
 
 On the character level, an LSTM with the dimension of 25 is used to embed tokens into a dense-space using character level training on each token. The results of the GloVe embedding and character embedding are concatenated as an input for the next layer. 
 
 In NeuroNER, bi-directional LSTM is used in the label prediction in the second layer. 

LSTM is a recurrent neural network architecture and is very effective in learning from sequence data like text. In the LSTM network in addition to input and output, each unit has three (input, output and forget) gates (although there are variations). With the input dimension $m$ and recurrent dimension $n$, the matrix of weights of input and recurrent connections have a dimension of $D_{W}=m*n$ and $D_{U}=n*n$, respectively. In addition to the gates, there is also the state of the unit. There are also biases ($b$) for each one and so, forward pass one has:
\[ 
i^{t} = \sigma(W_{i}x^{t} + U_{i}h^{t-1} + b_{i})  \]
\[
f^{t} = \sigma(W_{f}x^{t} + U_{f}h^{t-1} + b_{f}) \]
\[
o^{t} = \sigma(W_{o}x^{t} + U_{o}h^{t-1} + b_{o})  \]
\[
\tilde{c}^{t} = \tanh(W_{c}x^{t} + U_{c}h^{t-1} + b_{c})  \]
\[
c^{t} = f^{t} \circ c^{t-1} +i^{t} \circ \tilde{c}^{t} \]
\[
h^{t} = o^{t} \circ \tanh(c^{t})\]
where \(x^{t}(\in \mathbb{R}^{m}) \) is the input vector for the unit, \( i^{t}\) ,\( f^{t}\) and \( o^{t}\) are input, forget and output activation vectors , \( \sigma(x) = \frac{1}{1+\exp(-x)}\) and (\(\circ\)) is element-wise product. The number of parameters then can be calculated by $4(n*m + n^2 + n)$.
The uni-directional LSTM, where the sequence can be fed in when the $t$ value increases. One can make similar calculations when the sequence is trained in the decreasing order of $t$. In the former case, the network learns from what comes before a value, while in the later, the network learns from what is after a value. Bi-directional LSTM is utilizing two LSTMs to learn from all around the element. This is how the LSTM learns the sequential dependence between the words.

Conditional random field (CRF) is a powerful statistical method in machine learning. The conditional part refers to the fact that CRF is a family of conditional distributions with a structure. For sequence labeling, which is of interest in this work, linear-chain CRF is the relevant choice which has a linear structure. In this paper linear-chain CRF is called CRF in short. 

\subsection{Data sources}\label{data-sources}
The i2b2 challenge data set has been used in this work. The data contains a set of over 1300 patient records, which is the largest publicly available data set for de-identification. These reports contain (de-identified) protected health information (PHI) and it is divided into three sets; namely, training, validation, and test set. The training set contains more than five hundred reports. 

While the system is trained to label all different PHI types, the patient name is arguably most sensitive part and there are more than seven hundred patient surname as well as more than five hundred patient given-name occurrences in the training data where well over hundreds of them are unique. Here, the focus is on the patient names. 
\subsection{Statistical analysis and attack models}\label{Statistical analysis and attack model}
To investigate the sharing/leaking of sensitive data, we assume the adversary has access to almost complete information. Here the almost complete information simply means that the trained model as well as all the reports are available to the adversary except that the name which is altered. The goal here is to investigate the differences in $P$ from data sets that differ in only one part, names for instance.  If the outputs are distinguishable, it suggests there may be unintentional memorization or leakage of the last names in the training data. More specifically, new sets of data were produced from the original one. For the first data set type 1 or inside, names in the original data set were replaced with other names from the corpus. The next set, data set type 2 or outside, was constructed by the same procedure but the original names were replaced with the names from a dictionary of names that did not appear in the original set. Each of the inside and outside data sets was divided to three subsets, the one that only surname was replaced (SN), the one that only given name was replaced (GN) and the one that both surname and given name were replaced (GN $\&$ SN). The surname dictionary contains above 80000 names, while male and female dictionaries contain about 3000 and 5000 names, respectively. In order to make the comparison with the original training data, we also preserved punctuation and capitalization of the set and the names. For statistical attack, the model was trained on original data set and then used original, data set 1 (inside) and data set 2 (outside) for testing and compared their relative parameters (the output of softmax, to be precise). 

 The non-parametric Kolmogorov---Smirnov (KS) test \citep{razali2011power} was used to determine whether there is any difference in a data set and the reference probability distribution or between distributions of two data sets (and thus two-sample KS test). 
 This test gives access to statistic $D$ values. Statistic $D$ then can be used in numerical or counting algorithms to estimate/calculate the p-value. To calculate $D$, one has: \[
D_{m,n} = \underset{x}{max} |S_{m}(x) - S_{n}(x)|,  \]
which gives the maximum absolute difference between two distributions with m and n number of samples. The distributions are relative (normalized) and empirically produced from samples, and so, sometimes are called empirical distribution function.

Different re-identification attacks were attempted to stress-check the vulnerability of the model. They can be categorized to cut-off attacks and membership inference attack.
For the cut-off attacks, the goal was to identify a set of limits for different probabilities which can be used to re-identify the participants. In na\"{i}ve cut-off attack the goal was to find a limit that can differentiate between the original, data set (1) inside and data set 2 (outside). In brute-force cut-off attack the goal was to feed all possible names and find if the original name can be re-identified by a limit. For this purpose, another data preparation was done by randomly selecting three reports in which the patient name appears six times or more in the body of the report and then insulating the sentences containing the name (surname) and then calculating the $P$ for them with replacing the surname with all the over eighty thousand names in the dictionary. 

\begin{figure*}
    \centering
    \includegraphics[width=0.8\textwidth]{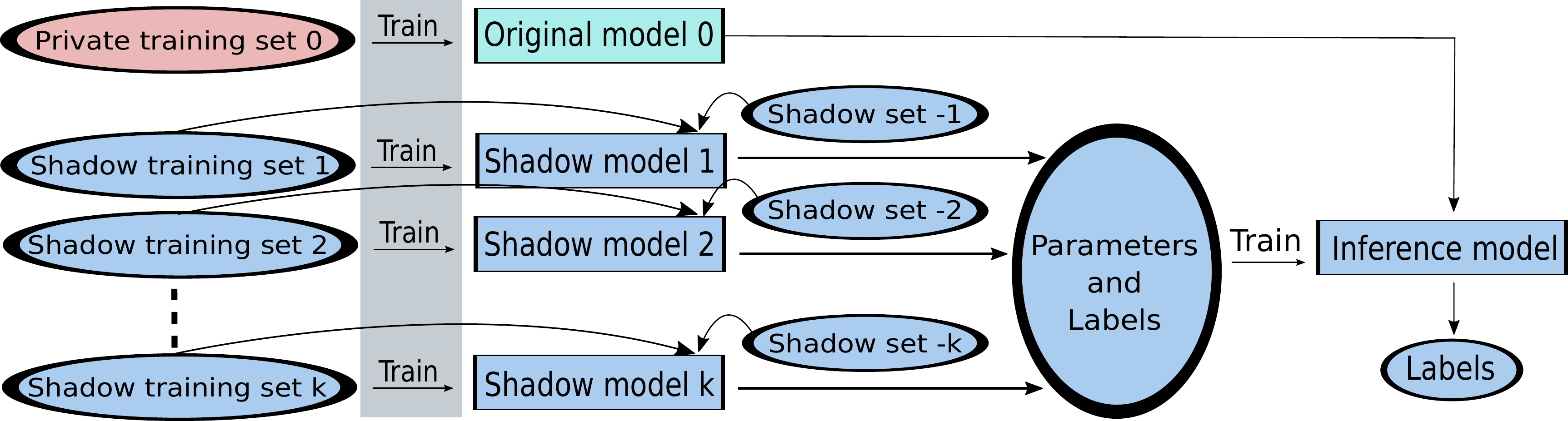}
    \caption{In this chart the ellipse represents data and rectangle indicates neural network model. The (-k) is the similar data where the names are replaced with other names that appears in all other sets but set k. Labels are indicating that the report is in the training data or not, and parameters are obtained from each model.}
    \label{fig:MIAFlow}
\end{figure*}

For membership inference attack, 12 different samples of type data set 2 (outside) were produced (shadow data sets \citep{shokri2017membership}), 10 of which were used to train 10 different shadow models. As figure
\ref{fig:MIAFlow} illustrates, shadow model 1 is trained on shadow data 1. Then the training data 1 along with a randomized mix of reports from shadow data 2 to shadow data 10 (data set -1) were fed as test to trained model 1 (the mixed data has the same number of reports as the shadow data 1). The $P$s then with label 1 for shadow data 1 and label -1 for mix shadow data 2 to shadow data 10 were extracted. The same process was used for all the 10 shadow models. These $P$s and labels then were fed in to a feed forward network to train it to label inside names (1) and outside names (-1). The attack accuracy is used as the metric. The shadow model 11 was used as validation set and shadow model 0 was the target model. The membership inference attack was conducted where the trained feed-forward network was used to differentiate and re-identify the names inside shadow data 0 using $P$s produced by testing the shadow model 0 on all possible names (brute-force) for a few reports with most repetition of a name.

\begin{figure*}
\centering
\includegraphics[width=0.8\textwidth]{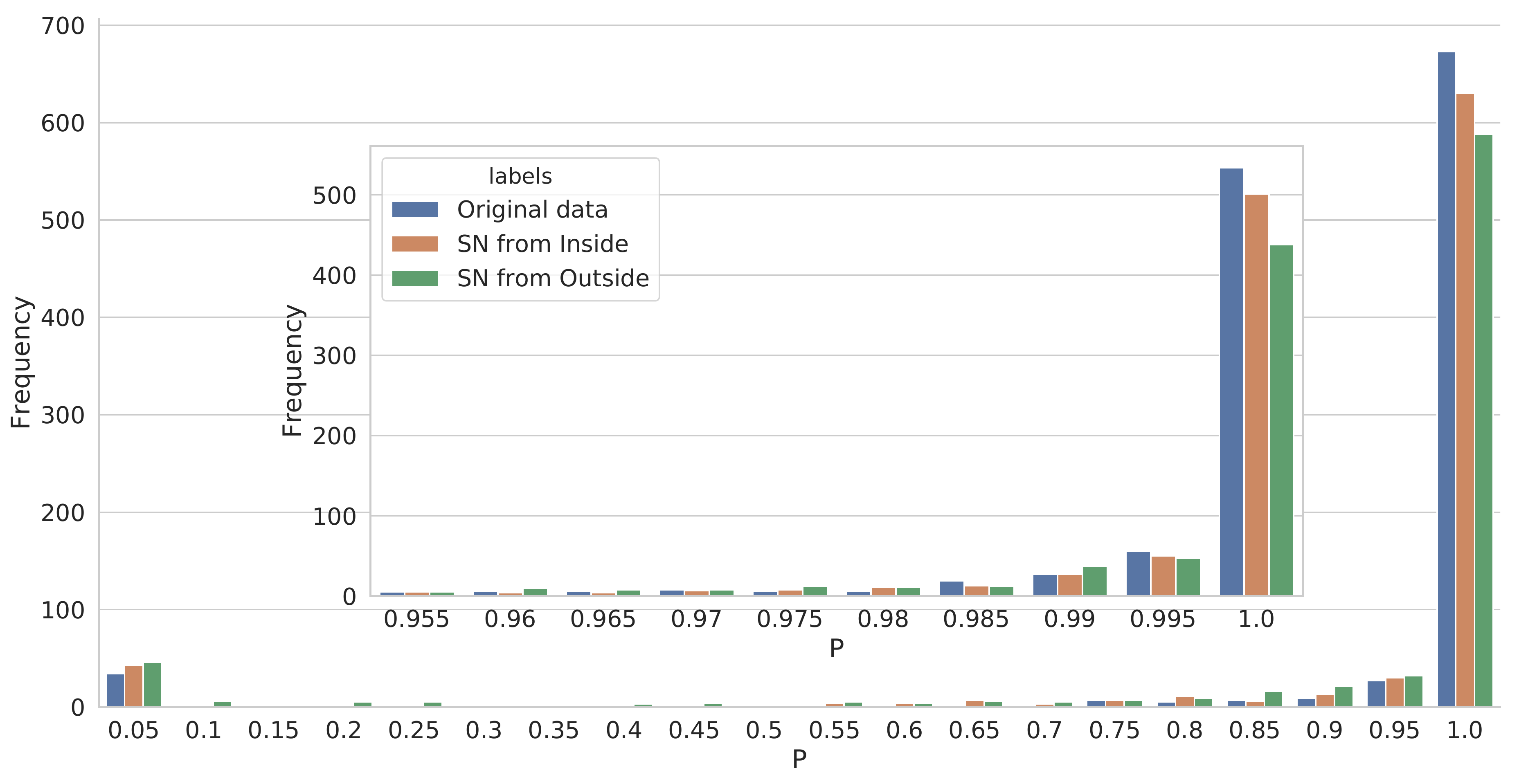}
  \caption{The graph shows the histogram frequency of probabilities for the case without CRF. The cyan represents the probabilities for original/unchanged data. The orange represents the probabilities of surnames being labeled names when altered randomly with other surnames from the corpus. The green gives the probabilities for surnames being labeled names when altered randomly with other surnames from the outside dictionary (excluding the in-corpus names).
  The inset zooms in the higher probabilities. The color-coding is the same as the main graph.}
  \label{fig:HPWOC}
\end{figure*}
\section{Parameters and Results}\label{results}
The software was trained on the data set with the following values for the  hyper-parameters:
Character embedding dimension and LSTM: 25;  Token embedding dimension and LSTM: 100; Optimizer: SGD; Learning rate: 0.01;Dropout rate: 0.5; Tokenizer: Spacy.

The precision achieved after 95 epoch is 98.43$\%$ on the test set and 99.96$\%$ on the training set. By dropping the third layer (CRF), after 88 epoch the achieved precision is 97.39$\%$ and 98.60$\%$ on the test and training data respectively. Note that the goal here is to investigate the risk of leaked information and not to necessarily achieve the best performance on the test set after training. Nonetheless, the high values of the precision, recall and F1 indicate that the system has been trained adequately (close to state-of-the-art) and is reflecting a real-world use of the algorithm. Also, the suggested precautions, such as freezing the token embedding, were implemented to maximize the implemented protection methods of the model. 



For the inference attack, a feed-forward network with one hidden layer of 64 ReLU units followed by a two softmax units provided the highest accuracy ($\sim$0.75). Please note the inference attack training data is balanced.

The magnitude of the test statistics $D$, and $p$-values obtained from the double-sided two-sample Kolmogorov-Smirnov tests were used to evaluate the difference in distributions as shown in table \ref{tab:ks2} for the network with CRF and the no-CRF network. As it can be seen, the null hypothesis that the original names give the same distribution as the altered ones does not hold, especially in the cases where outside data set 2 (outside) has been used (SN2 and SN2*). The question remains whether these differences are significant? And more importantly, can they be utilized for re-identification? Our cut-off attacks and membership inference attack does not indicate any potential for re-identification. The na\"{i}ve cut-off does not indicate any cut-off that can be used to re-identify any of the reports. The brute-force cut-off attack also could not narrow down the list of the potential original names to less than hundreds that would contain the original name used in the training. Even in the case of membership inference attack, the narrowest of lists got the original name ranked thirty-eight, while all the rest stayed well above hundred. In other words, these different attacks failed and no re-identification was possible. 

\begin{table*}
\centering
\begin{tabular}{||lllll||}
\hline
                           & \multicolumn{2}{c}{no CRF}                   & \multicolumn{2}{c||}{CRF} \\ \cline{2-5} 
\multicolumn{1}{||l|}{}     & D  & \multicolumn{1}{l|}{p-value}          & D   &  {p-value}          \\ 
[0.5ex] \hline
\multicolumn{1}{||l|}{SN1}  & 6.2e-2 & \multicolumn{1}{l|}{9.5e-2}         & 1.0e-1  & 4.3e-4         \\ \hline
\multicolumn{1}{||l|}{SN2}  & 1.6e-1 & \multicolumn{1}{l|}{6.0e-9}         & 2.3e-1  & \textless{}e-9 \\ \hline
\multicolumn{1}{||l|}{SN1*} & 6.8e-2 & \multicolumn{1}{l|}{5.6e-2}         & 9.1e-2  & 3.1e-3         \\ \hline
\multicolumn{1}{||l|}{SN2*} & 1.7e-1 & \multicolumn{1}{l|}{\textless{}e-9} & 2.3e-1  & \textless{}e-9 \\ \hline \hline
\end{tabular}
\caption{Two-sample Kolmogorov-Smirnov test $D$ statistic and double-sided p-value comparing the distribution of $P$s for surnames of original unperturbed data set and other sets which include: SN1 representing the surnames replaced from the corpus, SN2 surnames replaced from the outside dictionary, SN1* representing the surnames where both given-names and surnames have been replaced from the corpus, and SN2* representing surnames where the surnames and given-names where both replaced from the external dictionaries (exclusive of names in the corpus).}
\label{tab:ks2}
\end{table*}

\begin{figure*}
\centering
\includegraphics[width=0.8\textwidth]{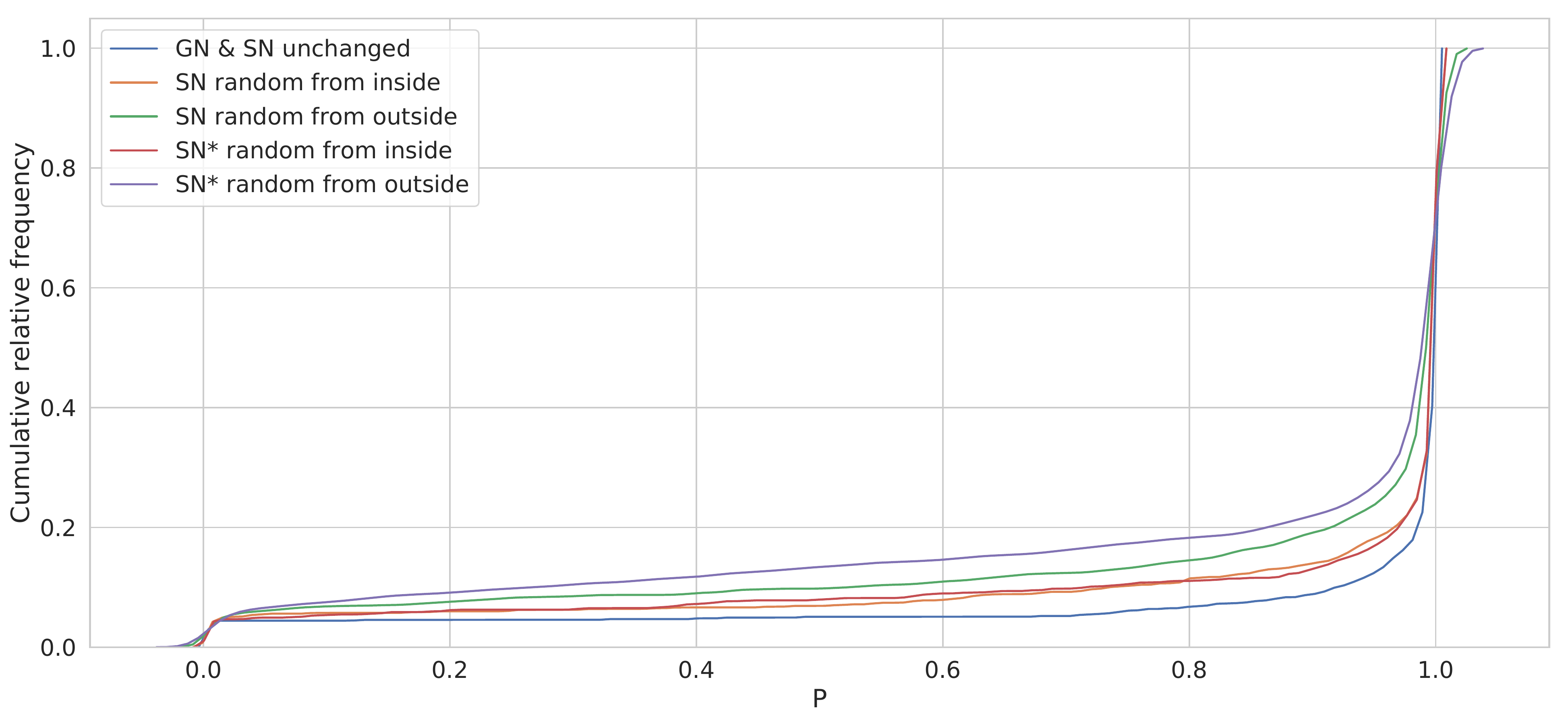}
  \caption{The kernel density estimates are provided for the values of $P$ for the network with no CRF. Cyan represents the unaltered original data, orange represents the case when surnames are replaced randomly with other names in the corpus, green represents the case when the surnames are replaced randomly with names from outside dictionary(exclusive), red represents $P$ values for surnames when both given-names and surnames have been replaced from other names in the corpus, and magenta represents values of $P$ obtained for surnames when both given-names and surnames were replaced randomly from the external dictionaries.}
  \label{fig:CHPWOC}
\end{figure*}

\begin{figure*}
\centering
\includegraphics[width=0.8\textwidth]{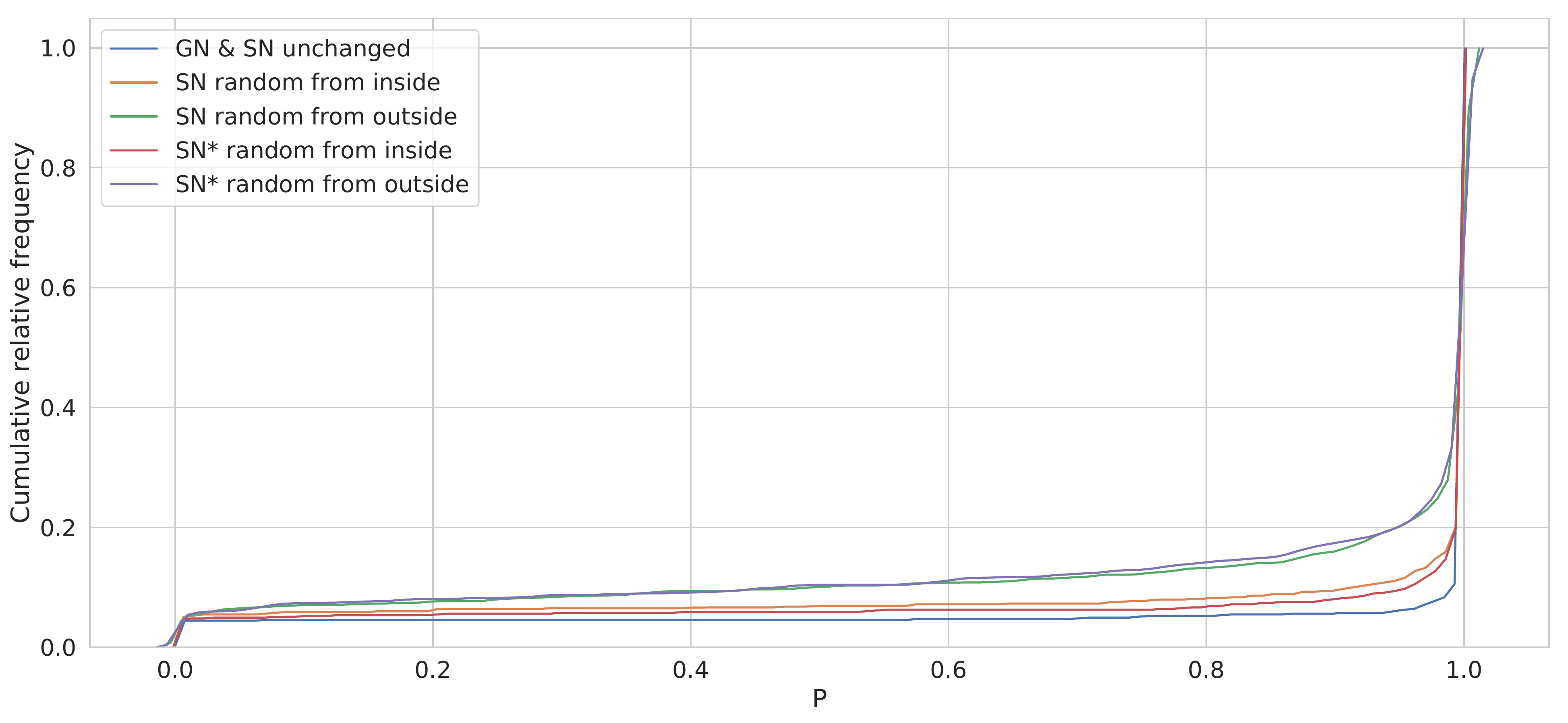}
  \caption{The kernel density estimates (curves) are provided for $P$ values for the network with CRF. Cyan represents the unaltered original data, orange represents the case when surnames are replaced randomly with other names in the corpus, green represents the case when the surnames are replaced randomly with names from outside dictionary(exclusive), red represents $P$ values for surnames when both given-names and surnames have been replaced from other names in the corpus, and magenta represents $P$ values obtained for surnames when both given-names and surnames were replaced randomly from the external dictionaries.}
  \label{fig:CHPWC}
\end{figure*}

The histogram (figure \ref{fig:HPWOC}) illustrates how the distribution of surname probabilities look like for the random replacement of the original names with the ones from other reports in the corpus, as well as from outside. The main graph shows the density of probabilities in different bins for different surname alterations. The inset figure illustrates the distribution of probabilities for the highest probabilities only. To the extent that there is a spread and widening in the statistics, they are overlapping. By using the cumulative distribution function, some differences can be more easily observed (figures \ref{fig:CHPWOC} and \ref{fig:CHPWC}). Figure \ref{fig:CHPWOC} shows the cumulative density with Gaussian kernel density estimates for the case when $P$ is extracted from a model without CRF. Figure \ref{fig:CHPWC} represents the case for $P$ calculated on a network with CRF. 
\section{Discussion}\label{discussion}

In the result section, the difference between distributions for $P$s has been discussed. These differences are more notable when comparing the original data set with ones replaced with the external dictionaries. However, one can not infer that any sensitive information has been compromised. More precisely, while the p-values are indeed small, one has to note that they have been gained by sampling over seven hundred surnames. Also, the differences are not by any means drastic. The overlap of the distributions is overwhelming, and that can protect sensitive data from adversaries' inference attacks. For instance, there is no cutoff that can be used to exclude original $P$s from either of the other samples. 

Moreover, the attempts to narrow down the number of names in the candidates list, by filtering names with high $P$s between the appearances of the same patient name in the same report, were also unsuccessful. This failure is due to the persistence of same names from outside of the corpus with high $P$s, and also with high fluctuations in the rank of the original name for its different appearances. Furthermore, it is notable that in this work the almost absolute knowledge is presumed, meaning that everything including punctuation, notations, and capitalization of the names were preserved in the replacements and altered data. Even the membership inference attack did not improve the case for the leakage of sensitive data as the attack was un-successful to re-identify any data. Also, the attack could not even limit the candidates for original name bellow several hundreds as was the case for the cut-off attacks. 
To check the robustness of the model, we implemented the membership inference attack on a reduced data set with fifty reports which is a tenth of the number of reports in the original set to push the model to over-train and potentially increase the chance of data leakage, but that did not lead to any leakage. This indicates all the more to the point that the model does not leak data when used responsibly. Please note we took suggested precautions as mentioned in the training section. 

It is worth mentioning that $P$ can be interpreted as the probability of different labels for any input. The distributions of these probabilities for the original names, as well as replaced names, have very similar properties like mean, median, standard deviation, and maximum value. That is the case both for when the model has been trained with CRF layer as well as when CRF layer has been disabled. In the case of using CRF, the interpretation of the results of applying softmax on the output of the forward neural network in the second layer as probabilities is not as accurate. But the numerical values and statistics of both with and without CRF models, follow each other closely and the discussions and conclusions provided, stand in both cases.

Differential privacy \citep{dwork2014algorithmic} \citep{abadi2016deep} has been gaining momentum in recent years. It is very reliable in the sense that when applicable, it can provide mathematical insurances to preserve the plausible deniability up to desired thresholds by introducing noise to the process in a controlled and measured way. To achieve this mathematically safe-guarded security, the parameters of this added noise should be set carefully depending on factors such as the number of epochs, the number of independent data samples and so on. While new tools like tensor-flow privacy library help with the implementation of differential privacy, there are caveats for the practical implementation of the algorithm in cases where the size of available data is limited. The performance of the algorithms trained in this manner suffers \citep{rahman2018membership} especially in cases with limited amount of data and complex models. Moreover, as the number of dependent inputs increases, the noise should increase, and so the performance gets even worse. Thus, while theoretically ideal, differential privacy may fall short of enabling protection of sensitive data by decreasing the accuracy of the model and so potentially increasing the probability of releasing sensitive data. Of course it would be interesting to see the above propositions investigated with actual implementation of differential privacy and its effects, but that is beyond the scope of this work.

In this paper we have presented an analysis of the potential for a state-of-the-art deep learning algorithm for de-identification to leak information concerning subjects' identities. We show that although statistically there is difference between the network reaction to inside and outside names, there is no evidence to suggest that a user could guess that any given subject was present in the training data, and that the deep neural network encoded the identities of the users in the training data in a way that creates a risk to the users. Of course, this does not preclude that some analysis sometime in the future might reveal the identity of a user, but one can always make the argument that future technology can do anything, and we feel that this is not a sufficient argument to be of concern about posting trained networks on medical data, even when the source training data explicitly contains identities of individuals. The legal framework governing PHI was developed to focus on portability (hence the `P' in HIPAA), and judge the trade-off between risk and benefit of sharing data. This should extend to a new generation of algorithms trained on such data.

\section{Acknowledgements}\label{acknowledgements}
GC, SN and SS are funded by the National Science Foundation, grant \# 1822378 
`Leveraging Heterogeneous Data Across International Borders in a Privacy Preserving Manner for Clinical Deep Learning'.
GC and LX are partially supported by the National Center for Advancing Translational Sciences of the National Institutes of Health under Award Number UL1TR002378.
The content is solely the responsibility of the authors and does not necessarily represent the official views of the National Institutes of Health. LX is partially supported by 
NIH R01GM118609 Decentralized Differentially-Private Methods for Dynamic Data Release and Analysis. SN is partially supported by the National Institutes of Health, award \# K01ES025445.

\bibliography{literature}
\end{document}